\documentclass[conference]{IEEEtran}
\IEEEoverridecommandlockouts
\usepackage{cite}
\usepackage{amsmath,amssymb,amsfonts}
\usepackage{algorithmic}
\usepackage{graphicx}
\usepackage{textcomp}
\usepackage{xcolor}
\usepackage{multirow}
\usepackage{booktabs}
\usepackage{subcaption}

\def\BibTeX{{\rm B\kern-.05em{\sc i\kern-.025em b}\kern-.08em
    T\kern-.1667em\lower.7ex\hbox{E}\kern-.125emX}}
\begin{document}

\title{Recursive Belief Vision Language Action Models\\

}

\author{
    \IEEEauthorblockN{ 
        Vaidehi Bagaria, Bijo Sebastian, Nirav Patel
    }
    \IEEEauthorblockA{
        \textit{Department of Engineering Design}, 
        \textit{Indian Institute of Technology, Madras}, 
        Chennai, India \\ 
        Email: {ed21b070@smail.iitm.ac.in}
    }
}

\maketitle
 
\begin{abstract}


Vision-language-action models must enable agents to execute long-horizon tasks under partial observability. However, most existing approaches remain observation-driven, relying on short context windows or repeated queries to vision-language models (VLMs). This leads to loss of task progress, action repetition under perceptual aliasing, and high inference latency. While semantic grounding is important, long-horizon manipulation fundamentally requires persistent, action-conditioned state representations. Current VLAs lack such representations and exhibit limited temporal and physical reasoning, making them ill-suited for multi-stage control.
This paper introduces RB-VLA, a belief-centric architecture trained with self-supervised world-model objectives that maintains a compact latent state encoding task-relevant history, dynamics and object interactions. Queried once per task, the VLM provides high-level intent, while the belief tracks task progress and enables phase-aware, causally grounded control under partial observability without storing raw observations or scaling memory with time. The belief and intent jointly condition a diffusion policy for robust closed-loop execution.
RB-VLA outperforms prior VLAs on long-horizon benchmarks, achieving 52.5\% and 37.5\% higher success rates on multi-stage pick-and-place and stacking tasks, respectively, compared to $\pi_0$. It also reduces inference latency by up to 5× relative to baselines and eliminates memory growth across timesteps observed in existing VLAs. The model demonstrates increased robustness under occlusion and more consistent task progression in multi-stage settings. Ablations show the belief module is the primary driver of performance, increasing success rates from 32.5\% without belief to 77.5\% with belief. These results demonstrate the effectiveness of belief-based state representations for long-horizon, multi-stage control.

\end{abstract}

\section{Introduction}
Vision–language–action models \cite{kim2024openvla,ye2025vla,zitkovich2023rt} have shown strong generalization by combining language understanding with visual perception to directly produce robot actions, improving scalability over classical control systems. However, most VLAs implicitly adopt a Markovian assumption, conditioning actions only on the current observation or a short history window \cite{jang2025contextvla,lin2025hif,wang2025lola}. Such reactive reasoning fails to capture long-term context under noisy observations, perceptual aliasing, occlusion and temporally static short observation histories. This limitation reduces its effectiveness for long-horizon, multi-stage, partially observable real-world manipulation. While longer context could address this, current VLAs scale memory through token accumulation, resulting in high latency and unreliable task state retention. Effective decision-making therefore requires reasoning over past actions and their consequences through compact state representations. 

Some approaches rely on defining multi-stage dynamics with explicit task-stage supervision for long-horizon behavior \cite{xu2025stare,ji2025robobrain}. However, manipulation is driven by smoothly evolving, action-grounded internal dynamics rather than discrete labels. Learning continuous dynamics better captures shared physical structure across tasks, whereas stage-based supervision is task-specific and poorly generalizable.

World models \cite{chen2022transdreamer} address this by modeling agent–environment dynamics, enabling future-aware planning in VLAs.
Although vision–language models excel at semantic inference, they lack explicit grounding in physical dynamics and state evolution, ignoring physical constraints \cite{gao2025vision}. World models bridge this gap by linking visual observations to action-conditioned dynamics that enable reasoning about future states beyond semantics alone. Existing approaches refine VLM decisions by predicting future observations with world models for online correction \cite{feng2025reflective,wang2025vagen,cen2025worldvla}, but rely on repeated, computationally expensive dynamic rollouts. In contrast, maintaining a compact internal belief state that tracks task progress through the consequences of past actions, enables action selection without continual forward simulation at each timestep. 

\begin{figure*}[t]
\centering
\includegraphics[
    width=\textwidth,
    height=8cm,
    keepaspectratio
]{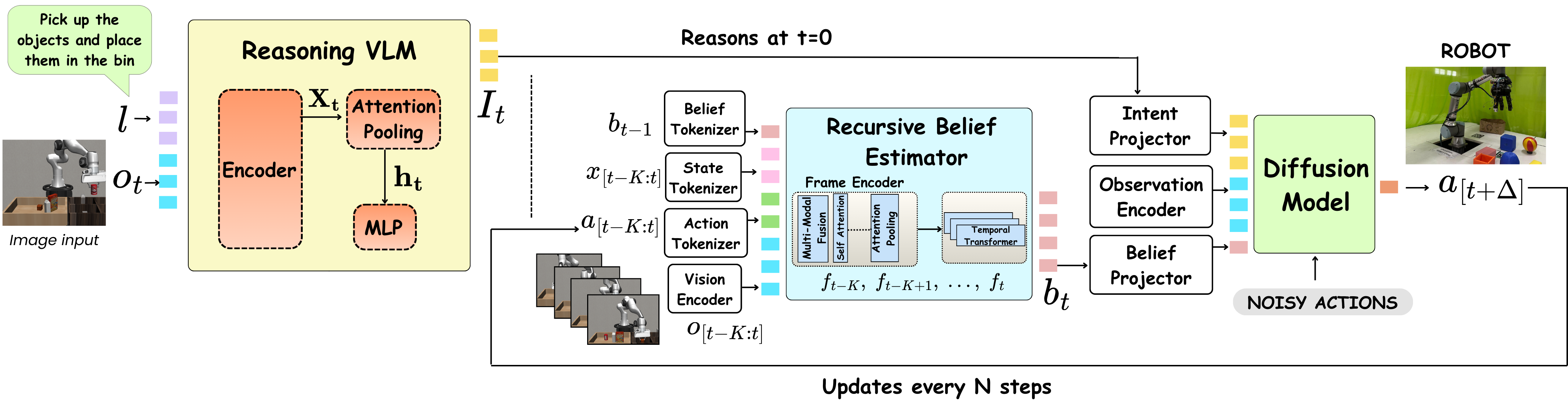}
\caption{\textbf{Overall Architecture of RB-VLA}. a) A VLM generates high-level intent from the current observation and instruction at t=0, implicitly grounding task semantics and relevant scene entities. b) Recursive belief estimator maintains a compact belief encoding task-relevant history. c) A DiT-based action head generates action chunks for closed-loop control}
\label{fig:block}
\end{figure*}

In the absence of an internal state, VLAs repeatedly re-interpret observations with VLMs, leading to high computational cost and latency. Since VLMs provide semantic abstraction, they can be invoked sparsely, while causal reasoning governs low-level task execution. Hierarchical approaches improve computational efficiency and robustness by decoupling planning and control for long-horizon tasks \cite{li2025towards,zhang2025hierarchical}. However, they still struggle with memory and task awareness.
To address these issues, we introduce the Recursive Belief VLA (RB-VLA) framework. Our key contributions are:
\newline
\begin{enumerate}
\item \textbf{Belief-centric VLA formulation:} We introduce a fixed-size, action conditioned recursive belief memory that captures task-relevant interaction history and system dynamics, enabling causal reasoning and consistent decision-making under partial observability.
\item \textbf{Phase-aware long-horizon control:} The recursive belief provides short and long-horizon predictive context, supporting implicit phase estimation and conditioning a diffusion policy for dynamics-consistent closed-loop control in place of short observation windows.
\item \textbf{Decoupled semantic grounding and control:} We treat semantic reasoning as episodic by querying a VLM once to extract grounded task intent, eliminating dense semantic re-inference. 
\item \textbf{Empirical validation on long-horizon benchmarks:} We demonstrate substantial improvements on long-horizon, multi-stage manipulation tasks and under occlusion, outperforming prior VLAs while maintaining constant memory usage and significantly lower latency.
\end{enumerate}

\section{Related Work}
\subsection{Vision--Language--Action Models}
Recent vision–language–action models \cite{kim2024openvla,brohan2022rt} remain reactive by framing control as autoregressive action token prediction using large language models. While effective for semantic alignment, invoking a full language model at every step to produce only a small number of action tokens is computationally expensive. 
Prior work attempts to reduce this cost through chunked or parallel decoding \cite{kim2025fine,song2025accelerating}, selective activation \cite{yang2025efficientvla}, and visual token compression \cite{vasu2025fastvlm}. However, discrete action token prediction limits smooth continuous control and precise multimodal action generation, motivating recent approaches that replace the language head with continuous action decoders, such as diffusion \cite{bjorck2025gr00t} or flow-based policies \cite{black2025pi_}. These approaches nevertheless remain fundamentally observation-driven, 
relying on dense semantic re-inference that recomputes the same goals and instructions at every step. 
RB-VLA addresses these limitations by maintaining a persistent state that tracks task progress and replaces computationally heavy VLM-based reinterpretation.
\subsection{Long Horizon Manipulation}
Long-horizon, multi-stage manipulation requires policies to reason over extended interactions. Existing approaches introduce inductive structure for long horizon tasks by explicitly predicting subgoals \cite{yang2025lohovla} or by conditioning policies on manually defined motion phases \cite{fan2025long}. While effective for skill chaining and temporal abstraction, these methods rely on task-specific subtask or phase supervision.

Some recent methods improve reasoning by introducing explicit intermediate inference before action selection. ECoT \cite{zawalski2024robotic}, ThinkAct \cite{huang2507thinkact}, and related approaches \cite{zhao2025cot,li2025coa} generate visually grounded reasoning over sub-tasks, affordances, or latent plans to condition action prediction. Although these methods improve reasoning and reduce hallucination, they operate primarily in image–text space and lack persistent awareness of task progress, often failing in multi-stage execution. This motivates belief-based dynamics-grounded representations that summarize interaction history and enable robust closed-loop long-horizon behavior.
\subsection{Memory and Context}
As interaction history grows unbounded over time, relying on noisy raw or compressed observation windows over long horizons becomes infeasible. Short histories discard task-critical temporal information, leading to failure under perceptual aliasing 
and in visually static multi-stage scenarios\cite{spaan2012partially}. In RB-VLA the belief instead provides a fixed-size, dynamics-consistent memory that filters observation noise and selectively retains motion-relevant information from past interactions and dynamics.

Some prior works incorporate history through recurrent features \cite{xiao2025ava} or growing memory tokens over time \cite{koo2025hamlet}, but these mechanisms do not explicitly ground memory in action-conditioned dynamics. Other approaches learn long-horizon memory via selective context retention and gated fusion \cite{liu2025evovla}, where memory primarily serves as a post-hoc modulator of policy outputs rather than an internal state that shapes perception and encodes action-induced transitions. Predictive world-model methods explicitly model dynamics and generate dense future rollouts for planning or inference \cite{cen2025rynnvla, liu2025trivla, lv2025f1}, while classical belief-space planners rely on explicit models and forward simulation \cite{kaelbling2013integrated}. Both incur substantial computational overhead and limit real-world applicability.

Using world-model training objectives, RB-VLA learns a belief state sufficient for future prediction, without heavy rollout-based forward simulation at inference. The belief recursively modulates perception toward interaction-driven dynamics. 
\section{Method}
\subsection{System Overview}
Long-horizon manipulation is formulated as a partially observable control problem with a learned belief estimator. The belief links perception to underlying motion dynamics rather than relying on raw observations, providing invariance to visual and environmental variation.

The system consists of three components (Fig.~\ref{fig:block}): a vision–language reasoning module, a belief estimator, and a diffusion-based controller. We decouple semantic task specification from causal state estimation: a vision–language model is queried sparsely to extract a static latent goal embedding $\mathbf{I}_t$ from language and initial observations, binding instructions to concrete objects in the scene. This grounded goal embedding remains fixed throughout execution, as object identity and task semantics are static.

At each step, the belief estimator integrates the previous belief $\mathbf{b}_{t-1}$, a window of tokenized observations $o_{t-K:t}$, actions $a_{t-K:t}$, and proprioceptive states $x_{t-K:t}$ (end-effector pose, velocity, acceleration and force–torque measurements) using a temporal transformer to produce a latent belief $\mathbf{b}_t$ that tracks physical interaction progress. Low-level control is executed by a diffusion policy conditioned on $(\mathbf{b}_t, \mathbf{I}_t)$, operating in closed loop at high frequency. 

\subsection{Belief Estimator}
\label{subsec:belief}
Long-horizon execution under partial observability is enabled by maintaining a belief state that captures information unavailable at a single timestep - such as occluded object states, contact history, and interaction progress—while being recursively updated and fixed in dimensionality. Similar to a state estimator \cite{chen2003bayesian}, this belief produces a coherent task representation for planning and control.

The belief transformer is trained as an RSSM-style world model \cite{chen2022transdreamer,chaudhry2026semantic}, separating deterministic dynamics from stochastic latent uncertainty. At each timestep, visual observations are encoded into patch-level features using a pretrained DINOv2 \cite{oquab2023dinov2} backbone, while low-dimensional robot proprioceptive signals are embedded into a shared latent space. These multimodal tokens are fused by a transformer-based frame encoder with multi-head self-attention and pooled via learned attention to produce a task-relevant frame summary. This frame summary $f_t$ highlights image regions that are linked to robot dynamics and produces interaction and dynamics-aware representations: 
\[
\alpha_i = \operatorname{softmax}\!\left(d_q^\top t_i\right), 
\qquad
f_t = \sum_i \alpha_i \, t_i
\]
where $t_i$ are the patch tokens and $d_q$ is a learned query vector.
We warm-start the frame encoder with a temporal contrastive objective that aligns nearby interaction states and separates dynamically inconsistent ones:
\[
s^+_{\delta}(t) = \langle f_t, f_{t+\delta} \rangle, 
\qquad
s^-_k(t) = \langle f_t, f_k \rangle,
\]
encouraging representations to reflect action-conditioned dynamics rather than appearance similarity.

Temporal integration is performed by a transformer over a recent fixed-length window of the frame encoder representations, providing short-term temporal context. To enable recursion across arbitrarily long horizons, the previous belief state $\mathbf{b}_{t-1}$ is pre-pended as a token, allowing the model to condition new evidence on accumulated dynamics-relevant history. This belief token actively conditions temporal attention, biasing integration toward relevant interactions and dynamics rather than simply propagating state. The corresponding output yields an evidence representation $\mathbf{e}_t$, which summarizes new observations conditioned on the prior belief. To capture uncertainty and branching futures, the model further introduces a stochastic latent variable $z_t$. The model learns a prior $p(z_t \mid \mathbf{b}_{t-1}) = \mathcal{N}(\mu_p, \sigma_p)$ representing predicted uncertainty before observing the next frame, and a posterior $q(z_t \mid \mathbf{b}_{t-1}, \mathbf{e}_t) = \mathcal{N}(\mu_q, \sigma_q)$ incorporating new evidence. The deterministic belief is then updated via,
\[
\mathbf{b}_t = \mathrm{GRU}\!\left(\mathbf{b}_{t-1}, [\mathbf{e}_t, z_t]\right).
\] 
The belief is trained using a standard ELBO objective on predicted future latent observations $x_{t+1}$ from an exponential moving average (EMA)-updated frame encoder, which stabilizes training and reduces target drift. Unlike prior methods \cite{chen2025vl} that predict future visual embeddings, our targets are action-grounded, causing the belief to capture dynamics rather than just appearance. In addition to one-step prediction, we include a lightweight multi-step auxiliary decoder that predicts $x_{t+5}$ encouraging the belief to encode longer-horizon dynamical evolution beyond immediate transitions. The reconstruction objective is:
\begin{equation}
\begin{aligned}
\mathcal{L}
=&\;
\mathbb{E}_{q(z_{1:T})}
\Bigg[
\sum_{t=1}^{T}
\Big(
\log p(x_{t+1} \mid \mathbf{b}_t, z_t)  \\
&
+ \lambda \log p(x_{t+5} \mid \mathbf{b}_t,z_t)
\Big)
\Bigg]  
\;-
\beta
\sum_{t=1}^{T}
\mathrm{KL}\!\left(q(z_t) \,\|\, p(z_t)\right).
\end{aligned}
\end{equation}
A KL divergence term with a scheduled weight encourages exploration early in training and structured belief refinement later. Introducing stochasticity prevents collapse to a single averaged future, allowing the belief to represent multiple consistent hypotheses and maintain a sharp, coherent latent state under partial observability.
To further enforce action grounding, we include an inverse dynamics loss,
\[
\mathcal{L}_{\text{inv}}
=
\left\lVert g(\mathbf{b}_t, \mathbf{b}_{t+1}) - a_t \right\rVert_2^2,
\]
encouraging the belief to encode controllable state information relevant for predicting executed actions.

The belief model runs at 10-20 Hz during inference and provides a persistent, anticipatory state sufficient for predicting future observations, enabling robust long-horizon execution without stage supervision or dense future rollout.

\subsection{High-Level Intent Prediction and Action Generation}
We adopt a vision–language–action framework in which a vision–language model (VLM) is decoupled with a diffusion-based low-level controller. The VLM is queried once per task using the instruction and the initial visual observation to produce an image-conditioned semantic task representation. This vector is projected through a lightweight MLP to obtain a compact intent embedding $\mathbf{I_t}$, which remains static throughout task execution, since the semantic goal is fixed for a given task.

The intent encodes the desired semantic outcome of the next execution phase - specifying \emph{what} should happen - while a diffusion policy determines \emph{how} it is achieved through continuous actions. During control, the intent is combined with the recursively updated belief state $\mathbf{b_t}$, enabling goal-directed control conditioned on the current execution stage.

Unlike approaches that repeatedly invoke a VLM for step-wise reasoning, we treat semantic grounding as an episodic process. The VLM is not re-queried during execution unless the task specification changes or replanning is required. Task progress is instead inferred implicitly through the evolution of the belief state, which tracks physical events such as contact, grasp, release, and object motion even when visual evidence is ambiguous or occluded.

\begin{figure}[h]
\centering
\includegraphics[width=8cm, height=5cm, keepaspectratio]{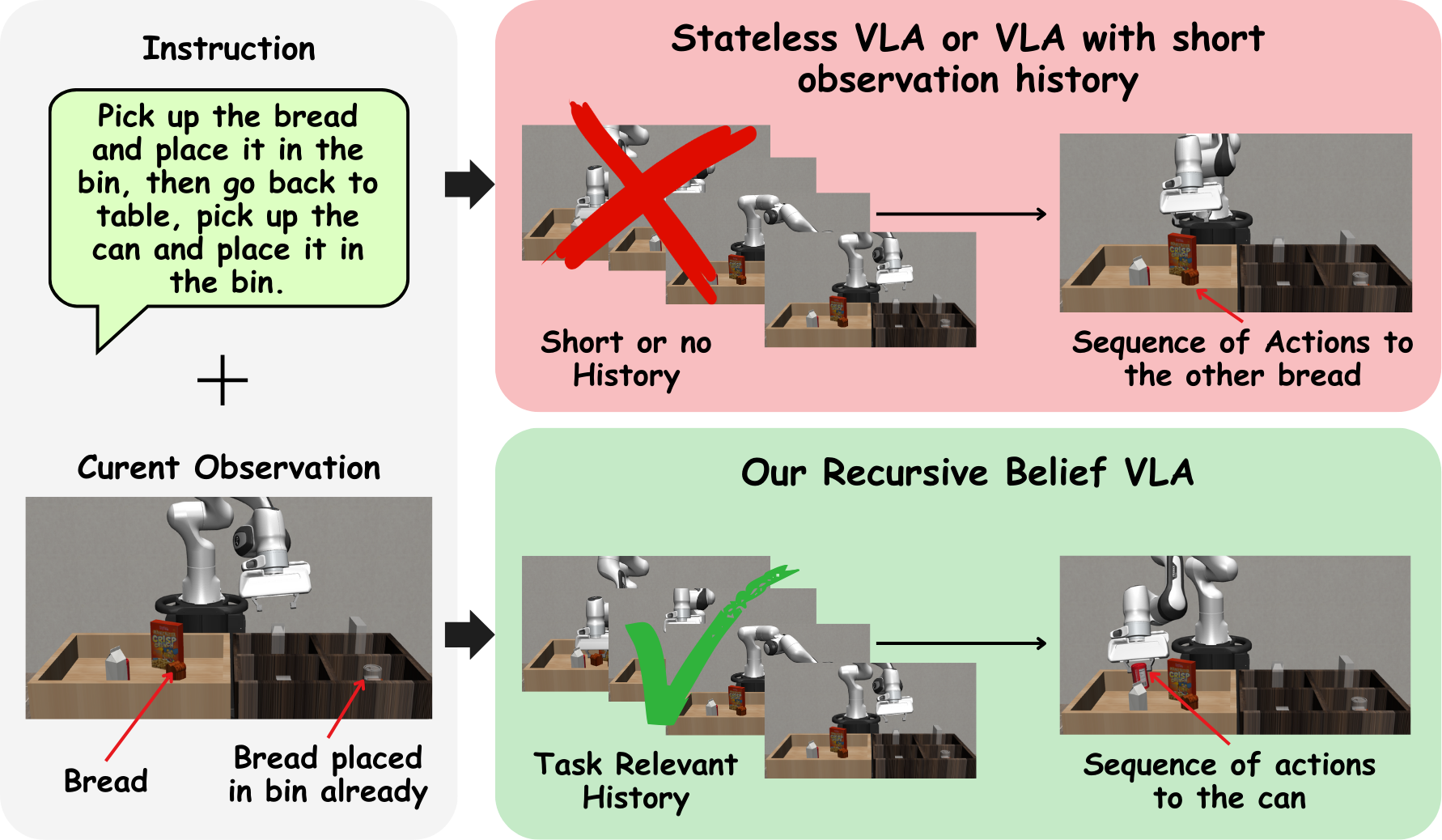}
\caption{Environment with similar objects}
\label{fig:block1}
\end{figure}
Fig.~\ref{fig:block} illustrates how the intent embedding is obtained from the VLM. We remove the language generation head and use the frozen vision–language encoder to extract the final multi-modal hidden states. Let $X \in \mathbb{R}^{(T+V)\times D}$ denote the final multimodal hidden states of the VLM after fusion, where $T$ and $V$ correspond to language and visual tokens respectively, and $D$ is the hidden dimensionality. To extract an image-conditioned task representation, we apply a single-query attention pooling over these tokens. Specifically, we introduce a learned goal query vector $\mathbf{q} \in \mathbb{R}^{D}$ and compute
\[
\alpha = \mathrm{softmax}\!\left(\frac{\mathbf{q}^\top W_k X^\top}{\sqrt{D}}\right),
\qquad
\mathbf{h}_{\text{task}} = \alpha X ,
\]
where $W_k$ is a learned projection matrix. This operation selectively aggregates visual and linguistic tokens into a single grounded representation. The pooled task vector is then projected through a lightweight MLP to obtain the intent vector $\mathbf{I}_t$. The MLP adapts high-dimensional VLM features to the control space. The intent extraction module is trained end-to-end with the diffusion policy using the action prediction loss, while the VLM backbone remains frozen. 
Figure~\ref{fig:block1} shows a multi-object manipulation task where short-horizon memory fails: once an object leaves view after being placed by the robot, stateless VLAs forget the interaction and reattempt grasping a similar object, while RB-VLA infers stage completion and continues the task.

\subsection{Low-Level Action Generation}
Low-level control is produced by a diffusion transformer policy that operates in latent space and is conditioned on the belief  $\mathbf{b}_t$, intent $\mathbf{I}_t$, and a fused state representation $\mathbf{s}_t$ from the state encoder.
This encoder fuses robot proprioception and DINOv2 image embeddings to the same dimensions, decoupling multimodal perception from belief- and intent-level reasoning. The policy samples action chunks from
\[
p\!\left(a_{t:t+H} \mid b_t, I_t, s_t \right),
\]
via iterative denoising, warm-started from the previously predicted action sequence with added low-variance Gaussian noise to improve sampling efficiency and temporal coherence. Actions are executed in a receding-horizon manner at high frequency (50 Hz), enabling smooth, closed-loop control under partial observability without reliance on dense semantic inference.

\section{Dataset and Training}
We use a staged training pipeline for  the vision–language module, the belief model, and the diffusion controller. The Belief transformer is first trained with the self-supervised objectives described in \ref{subsec:belief}. The intent extraction layers and the diffusion policy are trained jointly. All models are trained using mixed precision and gradient accumulation.

\subsection{Dataset}
Training uses 40,000 simulated manipulation trajectories ($\sim$40M timesteps) across 10 tasks, collected in RoboSuite, LIBERO-Long, LIBERO-Object and a UR5 MuJoCo environment. The dataset spans single and multi-object pick-and-place under occlusion, multi-block stacking in RoboSuite, and household tasks in LIBERO such as placing objects into drawers and cabinets. In practice, we observe that the belief converges reliably using a moderate number of trajectories, as the model focuses on learning control-relevant dynamics and action-conditioned state transitions rather than improving dense per-step semantic reasoning. We further finetune the belief and diffusion model with real-world manipulation tasks, adapting to sensor noise and unmodeled dynamics. During training, RGB observations are augmented with random brightness/contrast jitter, additive Gaussian noise, and minor spatial cropping to improve robustness and encourage learning of task-relevant state information.
\subsection{Training details}
The belief model uses a DINOv2 ViT-S/14 vision encoder and produces a 256-dimensional belief state from multimodal tokens. Temporal integration in the belief transformer uses a fixed window of $K = 5$ frames. The belief module contains 150M parameters and is trained for 80k iterations with batch size 128 and learning rate $1e^{-4}$. The VLM is initialized from Qwen2.5-VL-7B \cite{bai2025qwen2} and remains frozen during training. We replace the original language head with a lightweight intent extraction module, which is trained on top of the frozen VLM to produce a 256-dimensional continuous intent vector. The extracted intent, together with the 256-dimensional belief state, conditions the diffusion policy. The diffusion controller has 120M parameters and uses a DDPM objective with 100 steps to predict 16-step action chunks. It is trained for 100K iterations with batch size 32 and a learning rate 1e-4. All experiments are run on 12 NVIDIA A100 GPUs (80GB) using distributed data parallel training.

\section{Results}
Recursive Belief VLA (RB-VLA) is evaluated in simulation and the real world, using RoboSuite and LIBERO-Long. The evaluation focuses on: (a) Long-horizon tasks; (b)  Latency and memory growth; (c) Belief representation analysis; (d) Ablation studies; (e) Real world tasks.
\subsection{Long Horizon Tasks}
We evaluate RB-VLA on long-horizon, multi-stage manipulation tasks with and without occlusion, assessing robustness under partial observability as well as sustained task progression over extended horizons. Experiments are conducted in RoboSuite and LIBERO-Long environments. 

We ran 40 trials per task with different random seeds. Robustness was evaluated at inference by randomizing object identities and placements, environment configurations, lighting conditions, object order, as well as introducing dropped frames and observation noise. Performance is measured by task success rate, defined as completing the task within a fixed horizon without collisions or premature termination. Quantitative results for two-object pick and place under occlusion and stacking are reported in Table~\ref{tab:longhorizon}.
\begin{table}[h]
\centering
\caption{Long-horizon success rates (\%)}
\label{tab:longhorizon}
\resizebox{\columnwidth}{!}{%
\begin{tabular}{lcccc}
\hline
Method & Single P\&P & Multi P\&P & Single Stack & Multi Stack 
\\
\hline
GROOT-N1 & 52.5 & 20.0 & 65.0 & 32.5 \\
OpenVLA & 22.5 & $\times$ & $\times$ & $\times$ \\
$\pi_0$ & 55.0 & 25.0 & 67.5 & 42.5 \\
RB-VLA (ours) & \textbf{82.5} & \textbf{77.5} & \textbf{85.0} & \textbf{80.0} \\
\hline
\end{tabular}}
\end{table}
RB-VLA achieves substantially higher performance across all long-horizon tasks, exceeding $\pi_0$ by 52.5\% and GROOT-N1 by 57.5\% on multi-stage pick-and-place under partial observability, while OpenVLA fails to complete the task in this setting. These baselines repeatedly invoke VLMs for step-wise semantic reasoning, but lack awareness of task progression in multi-stage tasks and lose object or state awareness under occlusion. Belief-based models address this limitation by capturing temporal dynamics and enabling causal reasoning, whereas VLMs are primarily used for semantic inference. This demonstrates that belief-based state estimation is crucial for robust long-horizon task execution.

\begin{figure*}[t]
\centering
\includegraphics[
    width=\textwidth,
    height=8cm,
    keepaspectratio
]{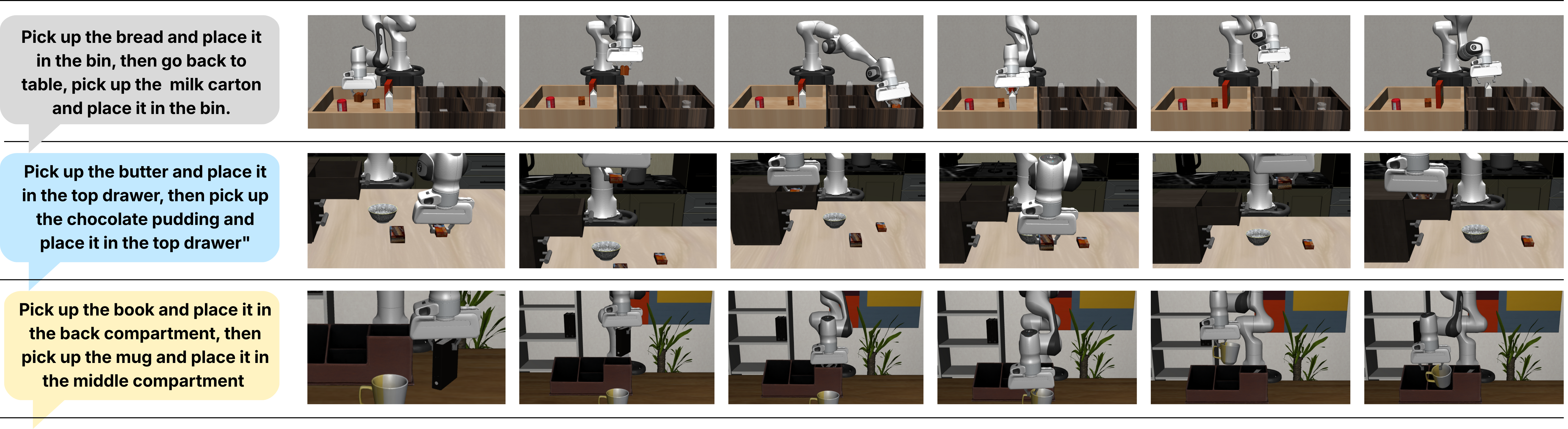}
\caption{\textbf{Long-horizon tasks under partial observability}. After the first object is placed and becomes occluded, the model correctly selects the subsequent target despite visually similar objects, avoiding re-grasping the same object.}
\label{fig:motion}
\end{figure*}

Figure \ref{fig:motion} illustrates representative multi-stage executions in which earlier object
interactions become occluded. The learned belief state preserves the necessary context to disambiguate visually similar objects and maintain consistent task progression.

\subsection{Latency and Memory Analysis}
\begin{figure}[h]
\centering
\includegraphics[width=8cm, height=30cm, keepaspectratio]{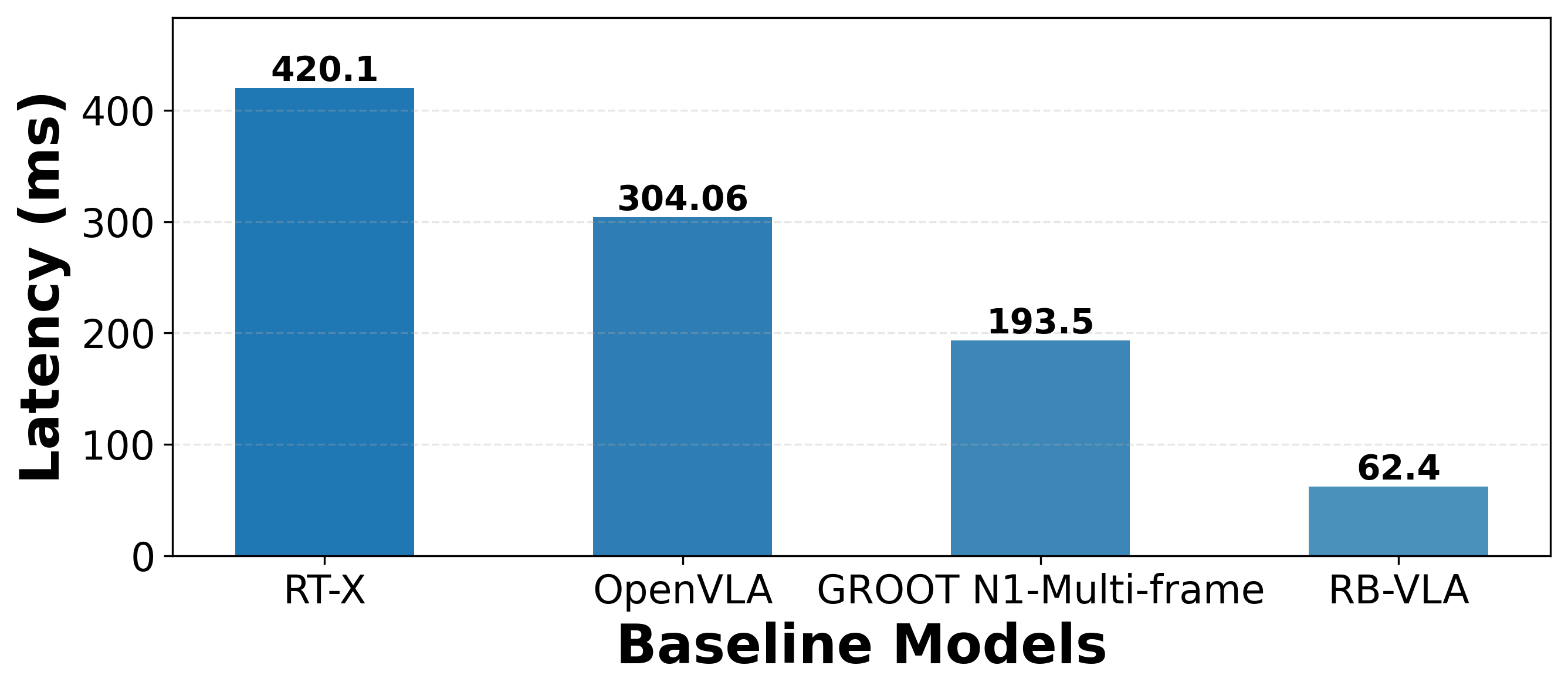}
\caption{Average inference latency per episode for RB-VLA and baseline VLA methods.}
\label{fig:latency}
\end{figure}

We evaluate computational efficiency using latency and memory, with all inference measurements performed on one NVIDIA A100 GPU. Figure \ref{fig:latency} compares the average episode latency of RB-VLA with prior methods. Since task instructions and semantic goals remain fixed within an episode, RB-VLA queries the VLM only once at the beginning of the task to obtain the intent embedding, and does not re-invoke it during execution. 
As a result, RB-VLA achieves over five times lower latency than stateless OpenVLA and RT1-X, and is more than three times faster than multi-frame GR00T-N1, which repeatedly processes multiple visual frames. This demonstrates that decoupling semantic grounding from control enables efficient real-time deployment. 

Many VLA systems maintain temporal context by accumulating multiple past frames or memory tokens, causing memory and compute to scale with interaction horizon. Table \ref{tab:memory_scaling} shows that both large VLM backbones (Qwen2-VL-7B) and end-to-end VLA systems (GR00T-N1) exhibit superlinear memory growth as the number of observed frames increases. In contrast, RB-VLA recursively compresses all task-relevant interaction history into a fixed-size belief state, yielding constant memory usage independent of episode length and enabling scalable long-horizon control.
\begin{table}[h]
\centering
\renewcommand{\arraystretch}{1.2}
\resizebox{\columnwidth}{!}{%
\begin{tabular}{lccc}
\toprule
\textbf{Temporal Context} & \textbf{Qwen2-VL-7B} & \textbf{GR00T-N1} & \textbf{RB-VLA (Ours)} \\
\midrule
1 frame  & $1.00\times$  & $1.00\times$  & $1.0\times$ \\
2 frames  & $1.81\times$ & $2.21\times$  & $1.0\times$ \\
4 frames  & $3.64\times$ & $3.85\times$ & $1.0\times$ \\
8 frames  & $6.82\times$  & $7.12\times$  & $1.0\times$ \\
\bottomrule
\end{tabular}%
}
\caption{Relative memory usage as a function of temporal context length, comparing RB-VLA with baseline models.}
\label{tab:memory_scaling}
\end{table}



\subsection{Belief Representation Analysis}
We evaluate belief quality by sampling pairs of belief states and testing whether proximity in belief space predicts similarity in future dynamics and control behavior. For each pair, we compute cosine similarity between their belief representations and compare it to (i) the $\ell_2$ distance between the corresponding future action sequences and (ii) the cosine similarity between their future visual feature representations. Unlike stage-labeled analysis with discrete phases (e.g., grasp, transport, place), manipulation dynamics evolve smoothly, thus belief quality is reflected in predictive consistency rather than phase separability.

\begin{figure}[t]
    \centering
    \begin{subfigure}{\columnwidth}
        \centering
        \includegraphics[height=0.18\textheight]{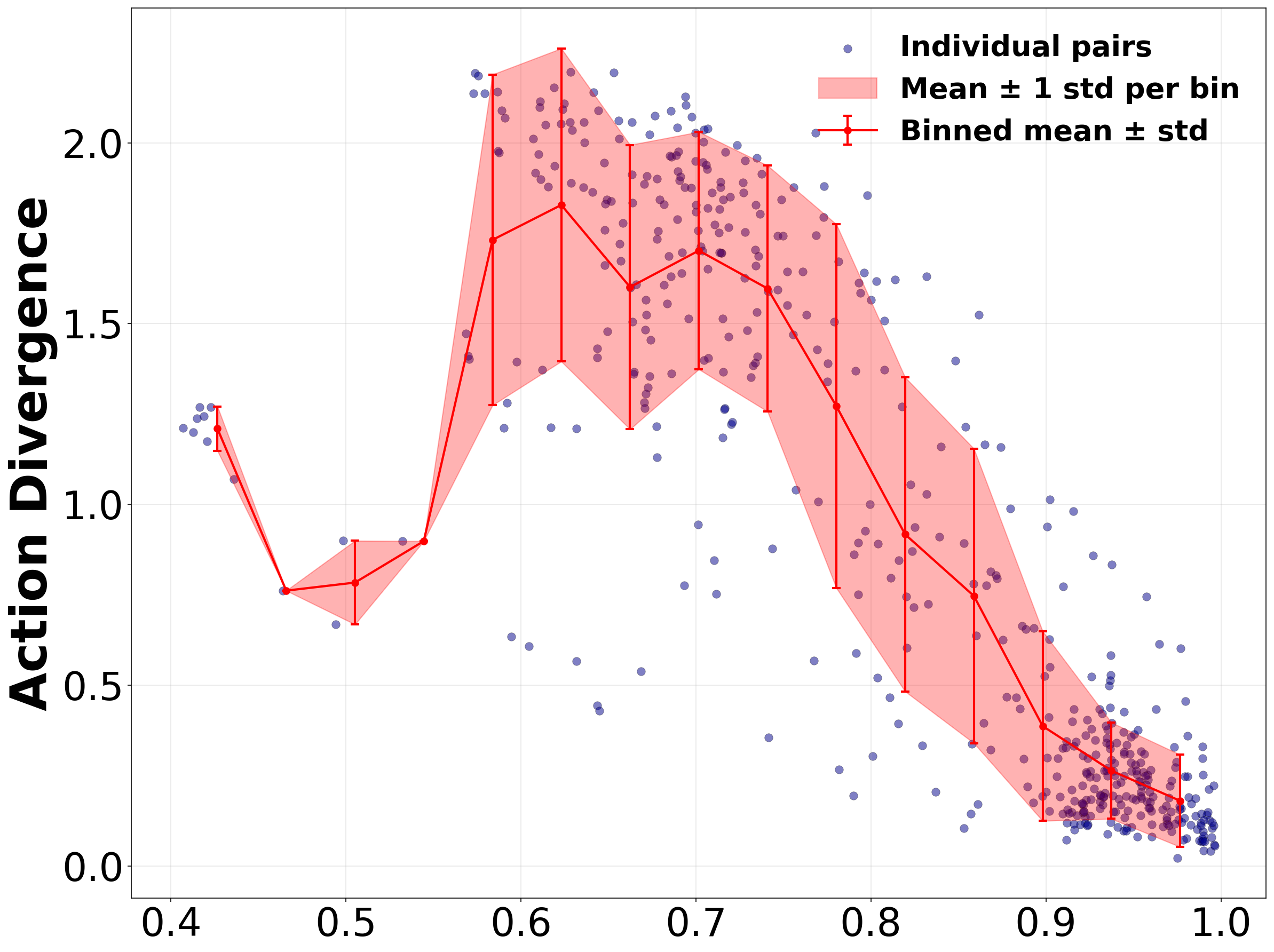}
    \end{subfigure}
    \vspace{4pt}
    \begin{subfigure}{\columnwidth}
        \centering
        \includegraphics[height=0.18\textheight]{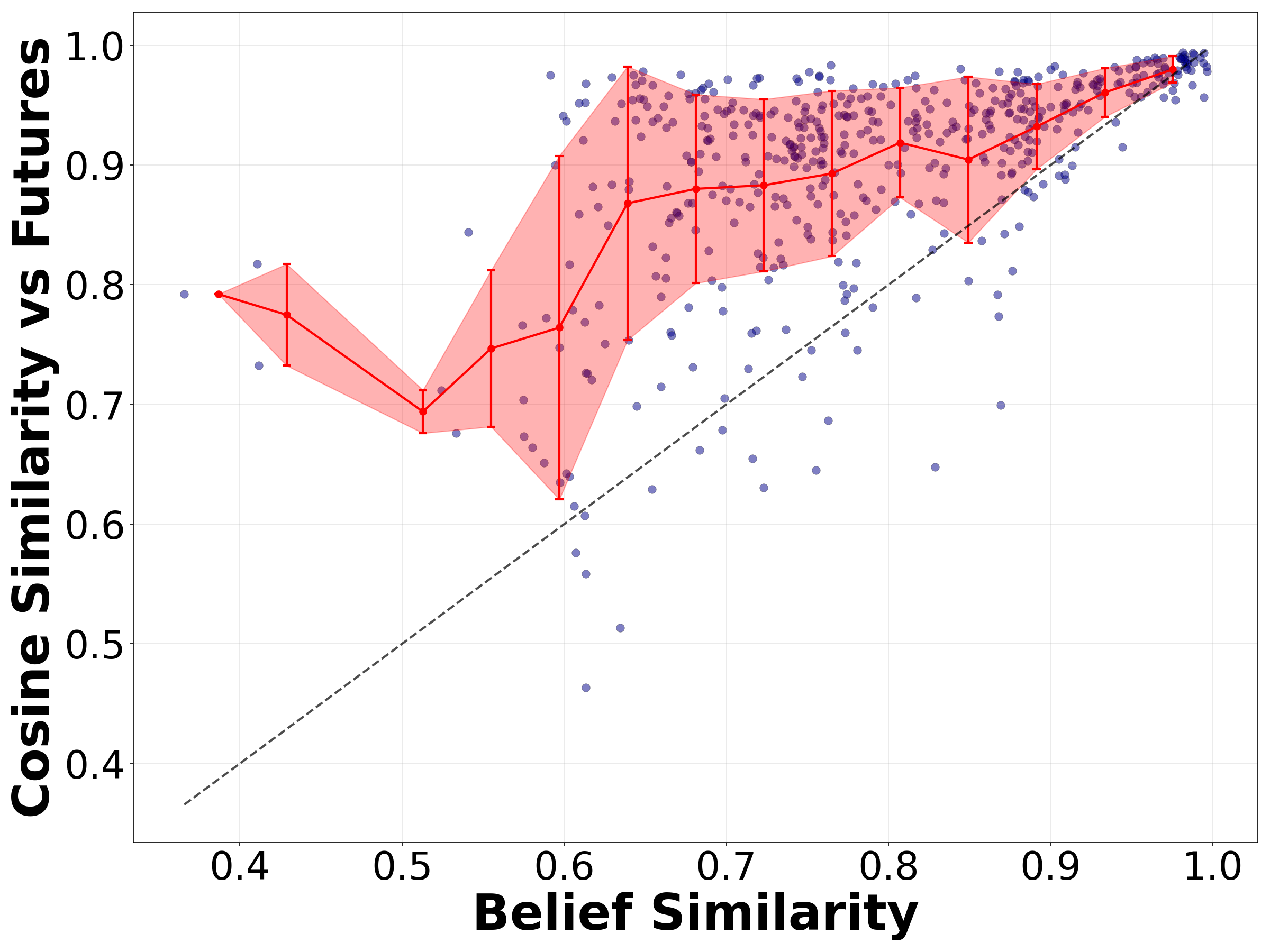}
    \end{subfigure}
    \caption{(a) Future $\ell_2$ action divergence versus belief cosine similarity. (b) Future observation cosine similarity  versus belief cosine similarity. Both plots exhibit strong monotonic correlation, indicating that proximity in belief space predicts similarity in future perception and control behavior. 
    }
    \label{fig:belief_similarity}
\end{figure}

As shown in Fig.~\ref{fig:belief_similarity}, belief similarity exhibits a strong monotonic relationship with both future visual similarity (Pearson $r=0.78$) and future action similarity ($r=0.87$). The shaded red region represents the variance, with $\sigma_{\text{bottom }20\%} = 2.3\,\sigma_{\text{top }20\%}$, indicating that similar beliefs encode consistent underlying dynamics and lead to similar action outcomes and future observations, while dissimilar beliefs correspond to different dynamical regimes, admitting multiple plausible future trajectories. This provides evidence that the belief acts as a latent action-conditioned dynamical state, governing future closed-loop evolution under control.

\begin{figure}[t]
    \centering
    \begin{subfigure}{\columnwidth}
        \centering
        \includegraphics[height=0.27\textheight]{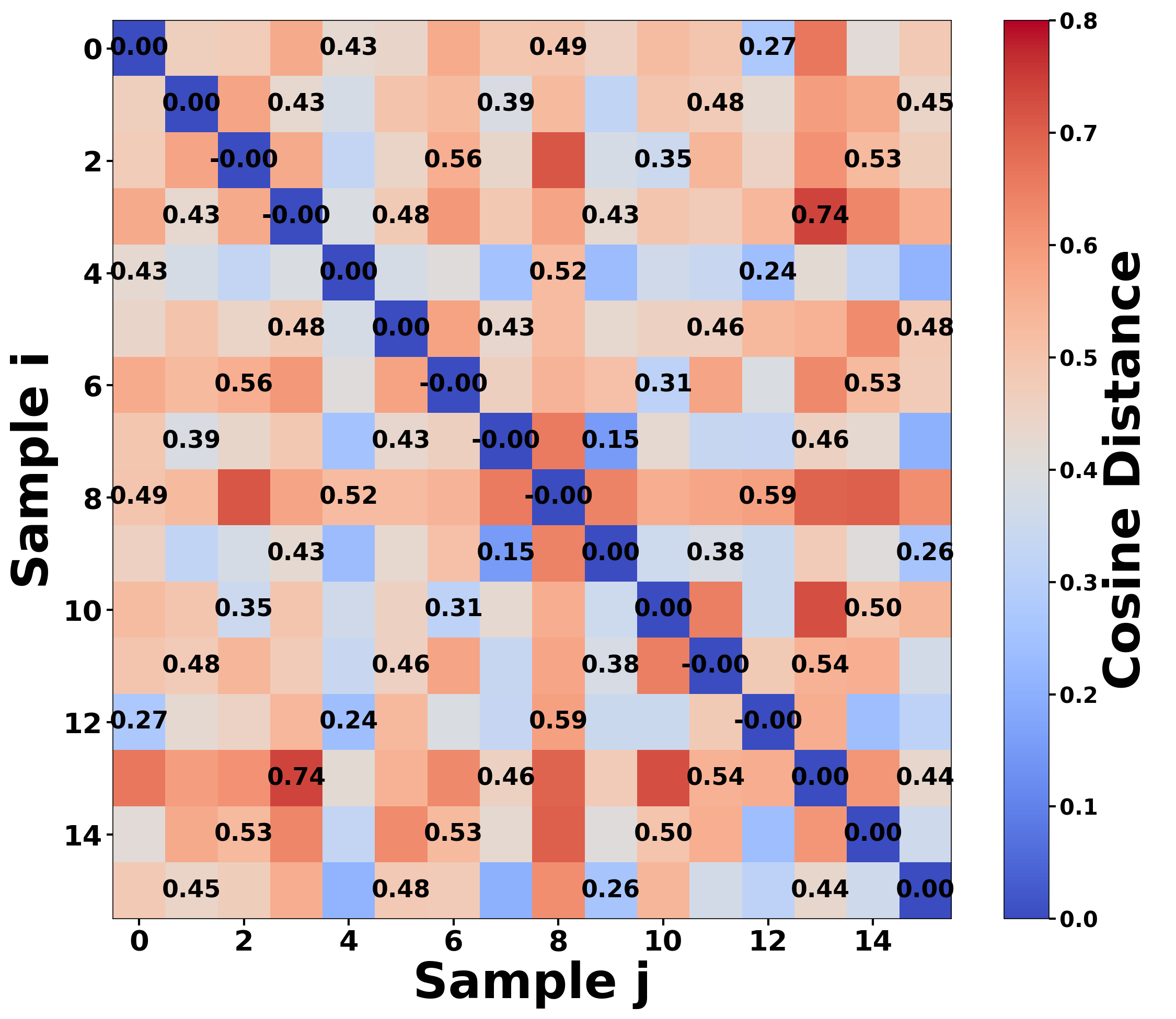}
    \end{subfigure}
    \vspace{8pt}
    \begin{subfigure}{\columnwidth}
        \centering
        \includegraphics[height=0.18\textheight]{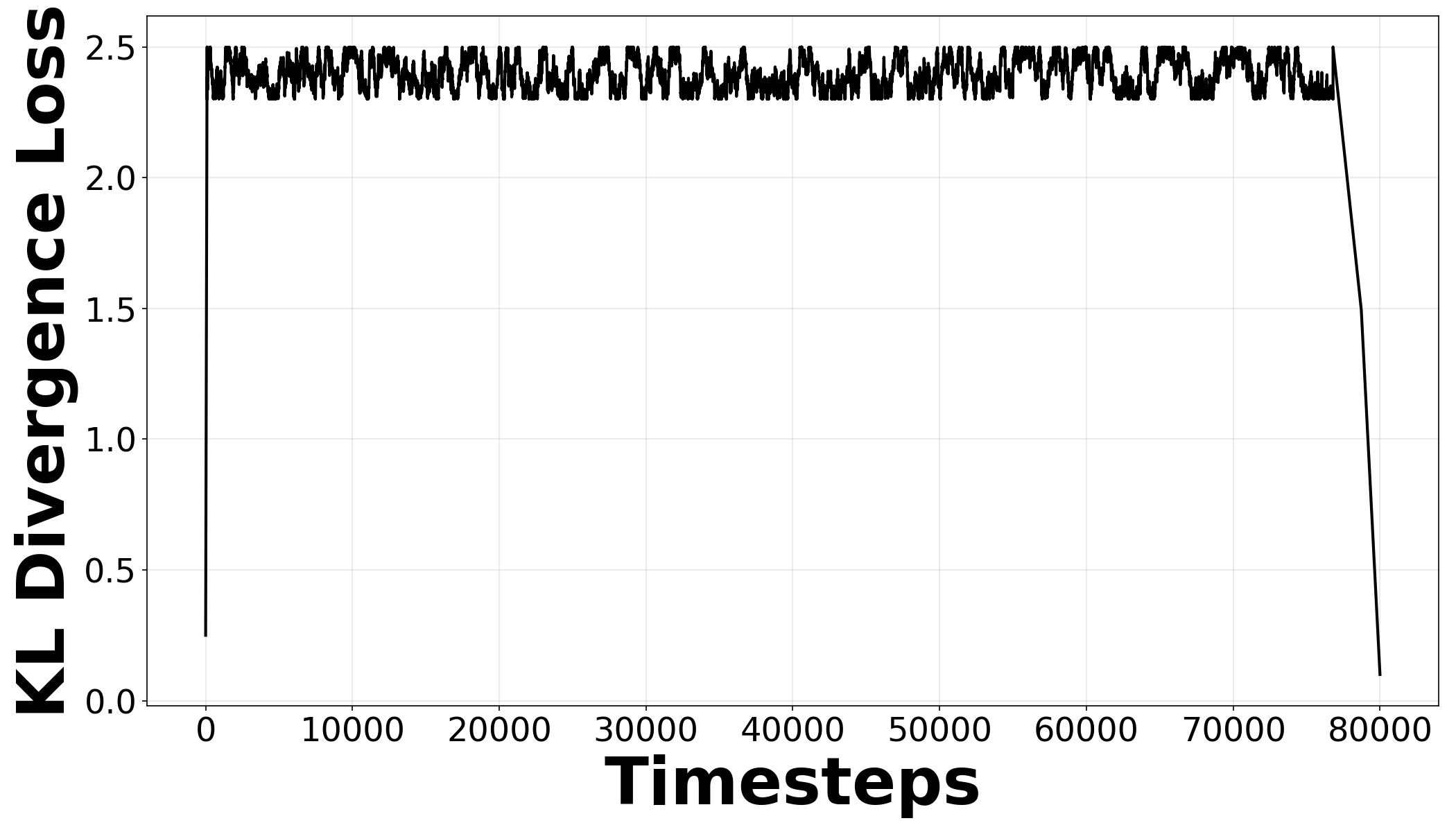}
    \end{subfigure}
    \caption{(i) Future divergence using cosine distance with multiple sampled latent variables ($n=16$) from a fixed belief state $b_t$, illustrating stochastic rollout diversity. (ii) KL divergence loss across training, showing stable convergence of the belief model’s latent distribution.}

    \label{fig:stochasticity}
\end{figure}

To analyze stochasticity, we fix a belief state $b_t$ and sample multiple latent variables $z_t \sim p(z_t \mid b_t)$. Figure ~\ref{fig:stochasticity} shows that different latent samples produce diverse yet structured future predictions, indicating that the belief encodes sharp underlying dynamics while $z_t$ captures uncertainty and multimodality over plausible future evolutions. The KL divergence stabilizes at a non-zero value, indicating sustained usage of $z_t$ and preventing collapse to a single averaged future.
These results demonstrate that the belief filters perceptual noise and preserves control-relevant structure.

\begin{figure}[h]
\centering
\includegraphics[width=8cm, height=30cm, keepaspectratio]{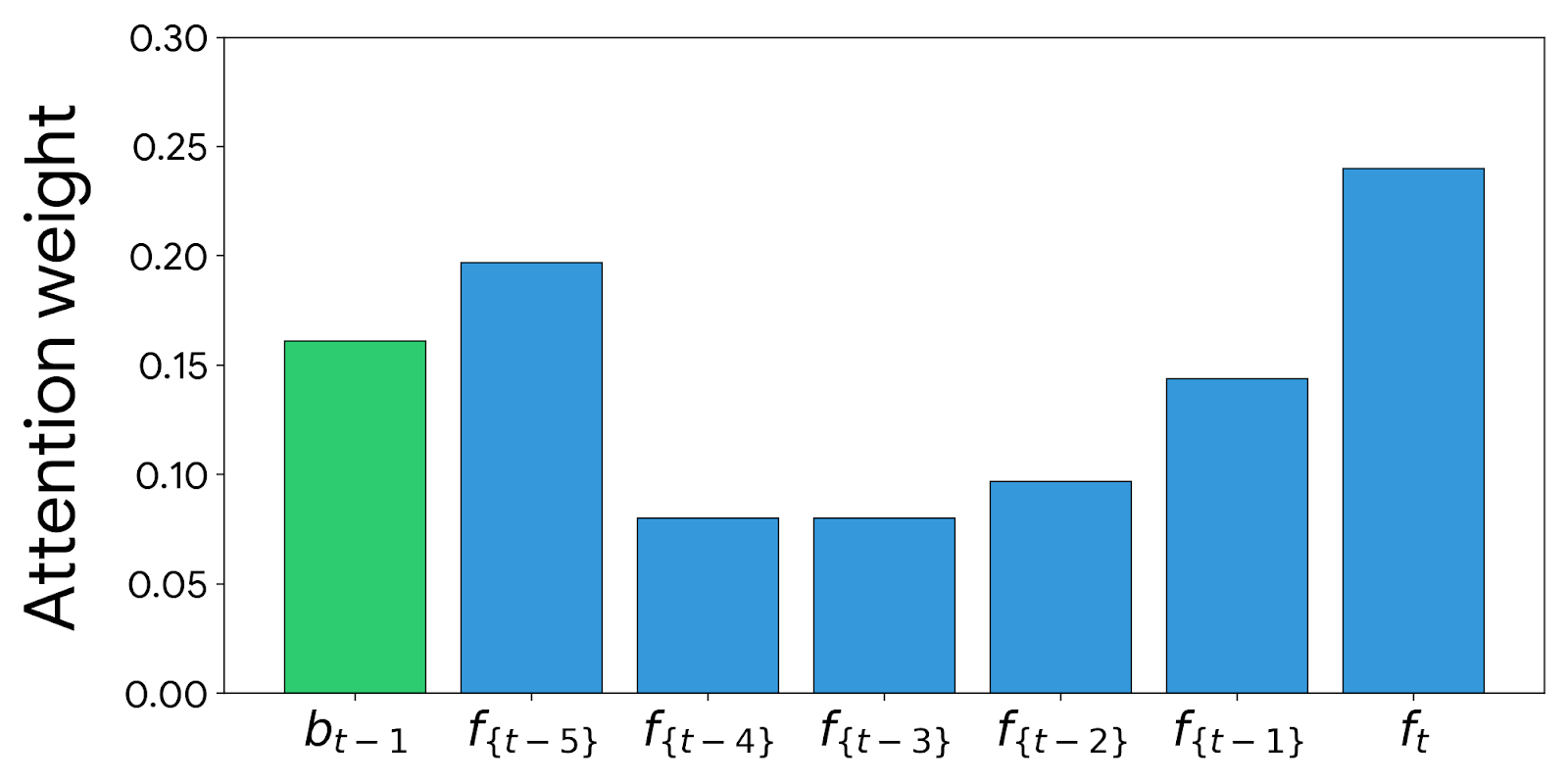}
\caption{Normalized self attention weights of the temporal transformer over the previous recursive belief token and short history of input frame encoder tokens, indicating which tokens contribute to the belief state at time $t$}
\label{fig:attention}
\end{figure}

Figure ~\ref{fig:attention} visualizes attention weights at timesteps where the manipulated object becomes occluded. The belief attends to its recursive state and causally informative past frames, adaptively retaining relevant dynamical history while discarding irrelevant information for future prediction.

\subsection{Ablation Study}
We ablate key components by selectively removing them and measuring the drop in average success rate over 40 episodes on the same long-horizon, multi-stage tasks under partial observability. Starting from the full RB-VLA, we evaluate: (i) no frame-encoder targets for training the belief (using only vision embeddings), (ii) deterministic belief without stochastic latent variables, and (iii) no belief–policy conditioning. As shown in Table \ref{tab:ablation}, each removal degrades performance, with the lowest performance when belief is decoupled from control (a standard DiT policy) and the highest success achieved by the full RB-VLA. This highlights the need for a dynamics-grounded belief state, since diffusion model alone lacks the dynamics-consistent causal state required for long-horizon, multi-stage behavior.
\begin{table}[t]
\centering
\caption{Component Ablation on Two-Object Pick-and-Place task. Average success rate over 40 episodes under partial observability.}
\label{tab:ablation}
\setlength{\tabcolsep}{6pt}
\renewcommand{\arraystretch}{1.0}
\scriptsize
\resizebox{\columnwidth}{!}{%
\begin{tabular}{cccc}
\toprule
\textbf{Frame Encoder} & \textbf{Stoch. $z$} & \textbf{Belief$\rightarrow$Policy} & \textbf{Success} \\
\midrule
$\times$ & $\times$ & $\times$ & 32.5 \\
$\times$ & $\times$ & $\checkmark$ & 57.5 \\
$\times$ & $\checkmark$ & $\checkmark$ & 62.5 \\
$\checkmark$ & $\checkmark$ & $\checkmark$ & \textbf{77.5} \\
\bottomrule
\end{tabular}%
}
\end{table}
\subsection{Real World Deployment}
We deploy RB-VLA on a physical UR5 manipulator under conditions closely matching the UR5 MuJoCo setup used in simulation. The model successfully transfers to real hardware without architectural changes, demonstrating effective sim-to-real transfer. The diffusion policy is further fine-tuned on 100 real-world trajectories to adapt to sensor noise and unmodeled dynamics. Using the learned belief representation, the system achieves reliable long-horizon execution on multi-object pick-and-place tasks under partial observability, maintaining low inference latency and stable closed-loop control despite visual noise and actuation variability. Over 25 real-world trials, RB-VLA achieves 68.0\% success rate, successfully completing grasp–transport–place cycles without manual intervention.
\begin{figure}[h]
\centering
\includegraphics[width=5cm, height=5cm]{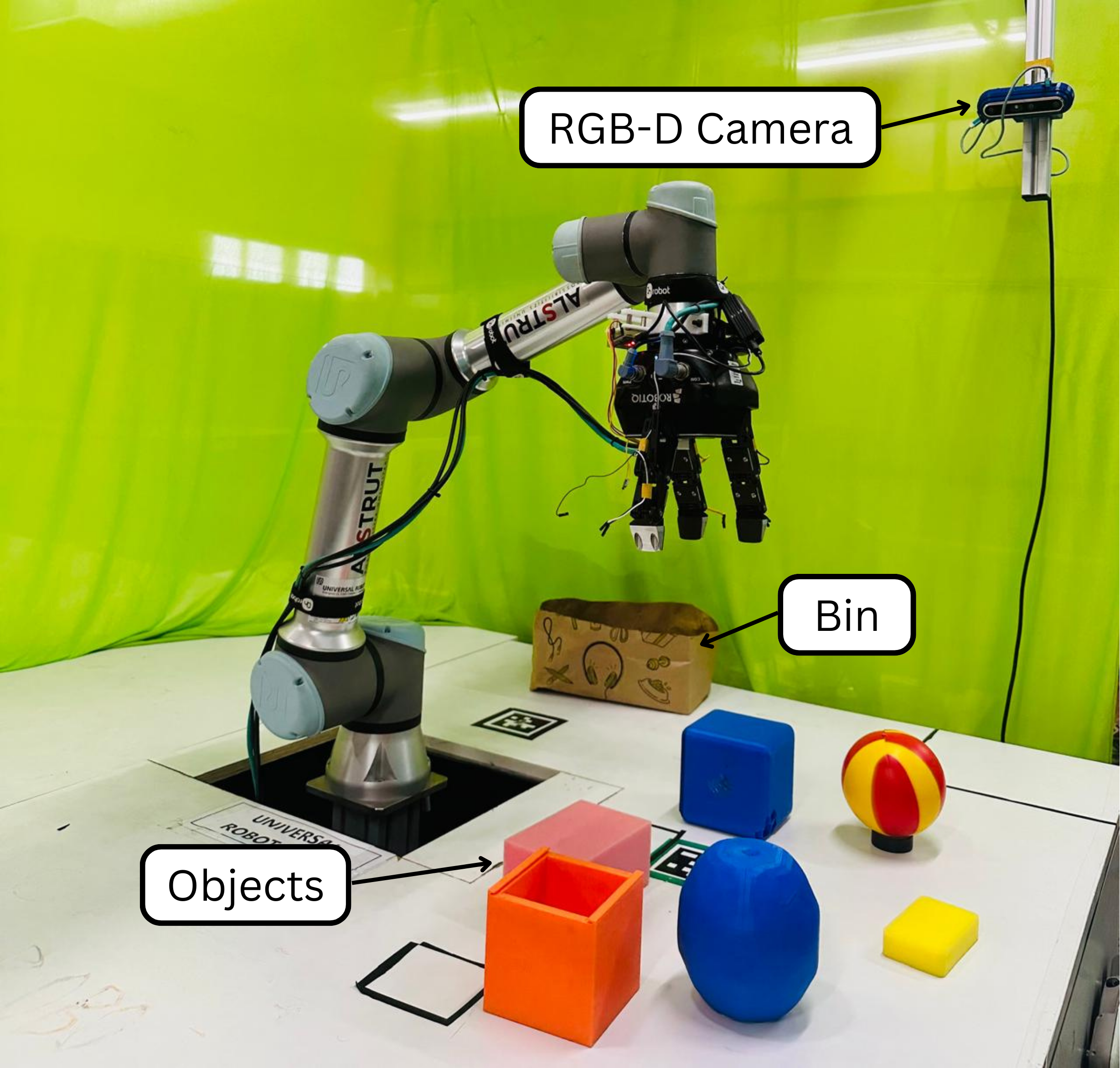}
\caption{Real World UR5 setup}
\label{fig:real}
\end{figure}
Figure \ref{fig:real} illustrates the real-world setup. An Intel RealSense D435i RGB camera is mounted to capture the gripper and workspace, and five objects of varying shapes and colors are placed in randomized configurations. Only RGB images are used for perception. After placement, the first object becomes fully occluded and is no longer visible in the camera view. Tasks are specified through natural language instructions describing the target object and goal location.

\section{Conclusion and Future Work}
We presented RB-VLA, a belief conditioned vision language action framework for long-horizon manipulation under partial observability. 
By maintaining a compact, action-conditioned belief state, the system preserves task progress and interaction history without growing memory, enabling robust multi-stage control with reduced latency. 
The architecture can be integrated into large pretrained VLAs as a fine-tuning layer for scalable robotic manipulation.

Future work will focus on scaling data and task diversity, integrating reinforcement learning for improved recovery behaviors, and extending the belief framework to support greater adaptability in dynamic environments.

\section{Acknowledgments}
We would like to thank Nikshep Grampurohit for his valuable contributions in refining the core ideas of this work. The authors also express great gratitude to Prof. Rengaswamy Jayaganthan for his constant support. 

\bibliographystyle{IEEEtran}
\bibliography{references.bib}
\end{document}